\documentclass{article}
\usepackage[preprint]{neurips_2026}

\usepackage[utf8]{inputenc}    
\usepackage[T1]{fontenc}       
\usepackage[hypertexnames=false]{hyperref} 
\usepackage{url}               
\usepackage{booktabs}          
\usepackage{array}
\usepackage{makecell} 
\usepackage{amsfonts}          
\usepackage{amsmath}
\usepackage{nicefrac}          
\usepackage{microtype}         
\usepackage{graphicx}
\usepackage{multirow}
\usepackage{xcolor}
\usepackage{float}

\title{Co-Evolving Harnesses and Models: On-Policy Correction Helps
       Weaker Models Catch Up Where Imitation Fails}

\author{%
  \textbf{Zhou Yu}\thanks{Corresponding author: \texttt{zhou.yu@salesforce.com}} \quad
  \textbf{Bin Bi} \quad
  \textbf{Shiva Kumar Pentyala} \quad
  \textbf{Shubham Mehrotra} \\
  \textbf{Sougata Chaudhuri} \quad
  \textbf{Shilpa Bhagavath} \quad
  \textbf{Zeyuan Chen} \quad
  \textbf{Ran Xu} \\
  \textbf{Phil Mui} \quad
  \textbf{James Zhu} \quad
  \textbf{Sitaram Asur} \\
  \normalfont Salesforce AI
}

\begin{document}

\maketitle

\begin{abstract}
Agent harnesses (the system prompt, tool set, execution hooks, and context-management scaffolding around a model) are a critical determinant of agentic task success, and automated harness evolution has recently proven highly effective at enabling smaller models to perform well on domain-specific tasks at a fraction of the cost of frontier models. Since both the harness and the model's weights shape an agent's behavior, and fine-tuning methods such as LoRA are a widely adopted practical model lever for small and mid-sized models, we ask how these two levers should be combined. Using seven enterprise agentic benchmarks \citep{yang2026harnesses}, we first evolve a harness with the weaker model and realize its gains; we then find that a stronger expert model often makes even better use of the evolved harness, suggesting that the weaker model could learn from the expert to close the remaining performance gap. However, the natural next step of teaching the weaker model using the expert's trajectories under the evolved harness surprisingly backfires: imitating the expert causes the weaker model to regress on all seven tasks ($-4$ to $-30$ points), a result reproduced across two model families (Qwen3-Coder and Gemma~4)\footnotemark[1], even though the identical procedure helps under the unevolved baseline harness. Our analysis reveals that, under imitation, the weaker model \emph{does} acquire the expert's knowledge and makes greater use of the harness scaffold, but its fit to the harness is impaired. The weaker model adopts the expert's planning strategy without the competence to execute it and no longer fits the harness that was evolved around its native planning style. To effectively introduce a teaching signal, we instead develop an on-policy expert-correction pipeline, automated end-to-end by a meta-level MLE agent: it localizes the failing turn in each of the weaker model's own rollouts and has the expert rewrite only that turn, preserving the model's planning style. This approach learns from the expert without breaking harness fit, thereby combining the benefits of both harness evolution and model adaptation. We identify and resolve a source of contention between harness evolution and model-weight adaptation, yielding a recipe can be incorporated into a co-evolution loop. Our results shed light on how to jointly and economically co-evolve harnesses and model weights to achieve strong performance on domain-specific enterprise tasks. 
\end{abstract}

\footnotetext[1]{\texttt{qwen3-coder-30b-a3b-instruct}
(30B total / 3B active) and \texttt{gemma-4-26b-a4b-it} (26B/4B active);
}

\section{Introduction}
\label{sec:introduction}

The capability of an agent is determined jointly by the underlying LLM and by
the harness that surrounds it---the system prompt, tool set, execution hooks,
and context-management scaffolding that determine what the model observes and
its action space. Recent work has shown that the harness is a powerful
lever, and treating it as an editable program and searching over it automatically
can lift a small open-weight model to near-frontier accuracy on coding and
enterprise tasks
\citep{yang2026harnesses,agrawal2025gepa,lee2026metaharness,harnessx2026}.
This is especially valuable for enterprise agentic tasks involving tool calling against internal APIs, code refactoring, and long-horizon data auditing, 
where inference cost also matters and a task-specific harness can provide the scaffolding needed for a smaller model to perform reliably at substantially lower cost.

Harness evolution, however, is only one of two levers for adapting an agent to a
domain; the other is the model weights. In this study, our interest is in the
co-evolution of the harness and model in the domain-specific regime, rather than
in broader post-training aimed at raising a model's general agentic ability.
Therefore, we mainly experiment with parameter-efficient fine-tuning, such as LoRA-SFT \citep{hu2022lora} which is widely
used to adapt a small or mid size model to a narrow task in a few GPU-hours. 
The central question we ask is: for task-specific domain adaptation under a reasonable budget,
how should harness evolution and lightweight model fine-tuning be combined, and 
can they be co-evolved to achieve synergistic gains rather than interfere with one another?

Expert trajectories are effective supervision for agentic model adaptation \citep{trajectory2task2026}, suggesting a natural next step after harness evolution. 
Prior work has shown that evolved harnesses transfer upward across
model backbones \citep{xu2026lifeharness}, and we confirm that a stronger expert
placed in a harness evolved for a weaker model adapts to it readily and often uses
it more effectively than the weaker model does. A natural training signal for
closing this model capability gap is therefore to imitate the stronger expert, and this suggests
a simple recipe: evolve the harness first, then fine-tune the weaker model under
the expert model's guidance. Surprisingly, we find that fine-tuning on expert
trajectories collected under the evolved harness consistently degrades task
success, even though the same recipe yields substantial gains under the
unevolved baseline harness, indicating that the failure arises from the
interaction between imitation and harness evolution rather than from imitation
itself. The regression is not a loss of harness usage or of knowledge; what
degrades is the fit between model and harness. The fine-tuned model copies the
expert's planning strategy without the competence to execute it, and no longer
matches a harness that was evolved around its native planning strategy.
Motivated by this diagnosis, we introduce the teaching signal on-policy instead:
a meta-level MLE agent localizes the failing turn in each of the weaker model's
own rollouts and has the expert rewrite only that turn, leaving the model's
planning distribution intact.

\begin{figure}[t]
    \centering
    \includegraphics[width=\linewidth]{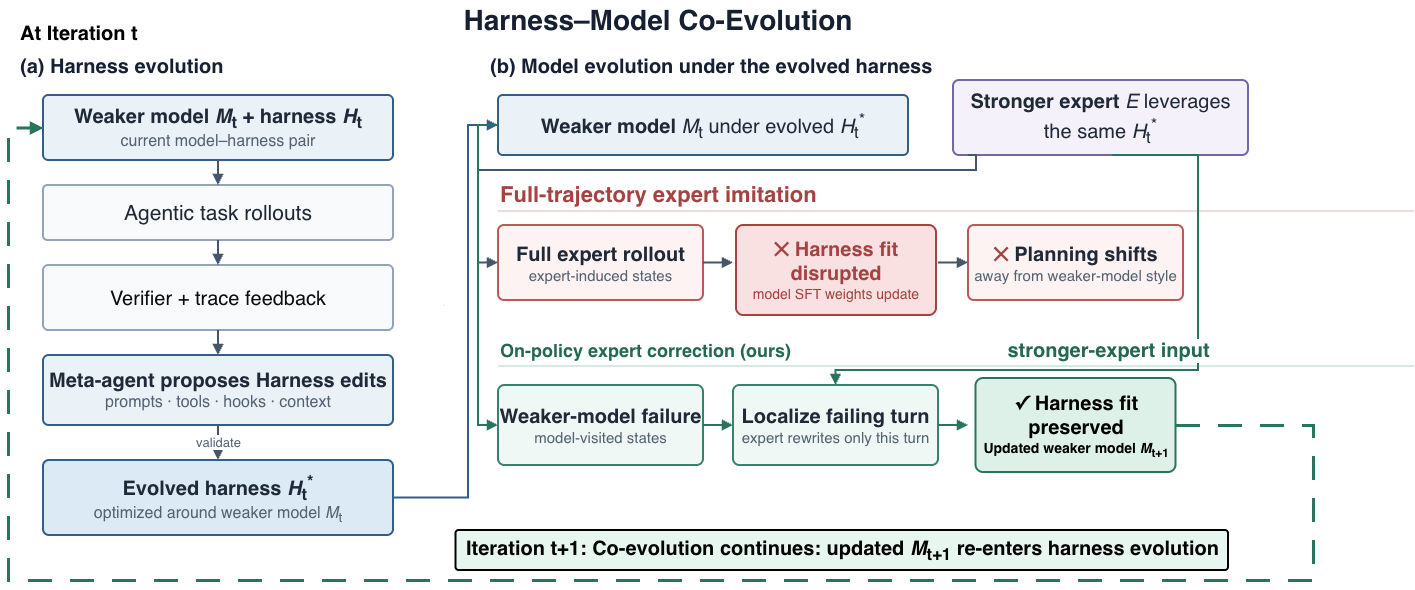}
    \caption{
        \textbf{Harness--model co-evolution.}
        Each round evolves a harness $H_t^{*}$ around the current weaker model
        $M_t$ and then updates the model under that harness.
        Full-trajectory expert imitation can disrupt the resulting
        model--harness fit, whereas on-policy expert correction uses the
        weaker model's visited states together with localized corrections from
        the stronger expert, producing an updated model $M_{t+1}$ while
        preserving the fit. The updated model can then re-enter harness
        evolution in the next round.
    }
    \label{fig:co-evolution}
\end{figure}

Our contributions are as follows:

\begin{enumerate}
    \item[(i)] \textbf{Upward harness transfer reveals model-side headroom and
    enables expert supervision.}
    Across seven enterprise agent tasks, a stronger expert readily operates the
    harness evolved for the weaker model and uses its task-specific adaptations
    effectively. Under this shared harness, the expert outperforms the weaker
    model on every task and matches or exceeds its own default-harness
    performance, demonstrating that the evolved harness remains useful across
    model backbones while exposing capability that the weaker model may not yet
    have realized.

    \item[(ii)] \textbf{Full-trajectory expert imitation breaks
    model--harness fit.}
    After harness evolution, fine-tuning the weaker model on trajectories from
    the expert model degrades task success on all seven benchmarks by 4 to 30
    points, and the effect persists with a second model family, while the same
    procedure yields gains under the unevolved harness, isolating the failure to
    the interaction between imitation and harness evolution. The cause is not lost
    knowledge or scaffold usage (both increase after SFT), but planning-strategy
    drift: the model adopts the expert's strategy and no longer matches a harness
    evolved around its own.

    \item[(iii)] \textbf{On-policy expert correction resolves this conflict and
    enables iterative co-evolution.}
    We develop an on-policy expert correction loop to mitigate the planning regression caused by direct imitation. A 
    meta-level MLE agent localizes the failing turn in the model's own rollouts
    and has the expert model demonstrate only that turn. We show that this approach can achieve further model-side gains on top of the gain from the evolved-harness, and can be incorporated into a harness-model co-evolution loop (Figure~\ref{fig:co-evolution}).
\end{enumerate}

More broadly, our results indicate that harness and model adaptation cease to be
independent once a harness has been optimized for a particular model: the
resulting scaffold becomes coupled to the model's planning and execution
behavior, so subsequent weight updates should preserve rather than inadvertently
disrupt that fit. For practitioners building cost-effective enterprise agents, this suggests a concrete design principle: 
after harness evolution, provide on-policy supervision at states visited by the student model instead of relying on wholesale imitation of a stronger model's trajectories. 
This allows model updates to build on harness improvements and supports iterative co-evolution.

\section{Related Work}
\label{sec:related}
Prior work has shown that optimizing over individual prompts or broader harness implementations can enable smaller models to perform complex, multi-step, tool-using agentic tasks \citep{yang2026harnesses,agrawal2025gepa,lee2026metaharness}. However, the boundary between harness and model adaptation is porous: behaviors introduced through external scaffolding can subsequently be absorbed into model weights \citep{dennis2026compiling,lu2026skillzero}. Recent work has therefore begun to optimize both levers. SIA interleaves harness and model updates on classification and scientific optimization tasks; Co-Harness fine-tunes a model on its own verified successful trajectories generated under an improved harness for mathematical reasoning; and HarnessForge and HarnessX jointly optimize model policies and evolving harnesses \citep{sia2026,chen2026coharness,harnessforge2026,harnessx2026}. Although these studies report gains in their respective domains, it remains unclear how the two levers should be coordinated once a harness has become specialized to a particular model, and under what conditions subsequent weight updates will compound rather than disrupt that fit. We investigate this question in heterogeneous, long-horizon enterprise agent tasks, focusing specifically on supervision generated by a stronger model as an additional teaching signal.
Agent post-training commonly uses supervised fine-tuning on successful expert trajectories, which can substantially improve tool use and task completion \citep{trajectory2task2026}. Recent approaches move beyond wholesale trajectory imitation by obtaining teacher continuations or interventions at states induced by the student policy \citep{lauffer2025oec,ye2026teachersteps}. However, these methods adapt the model under a fixed agent interface and do not examine how their supervision interacts with an evolved, model-specific harness. Our work connects these lines by showing that imitating complete expert trajectories can break a weaker model's fit to such a harness, whereas localized expert correction at states visited by the weaker model preserves this fit and allows harness and model adaptation to compose, providing a practical principle for their co-evolution.

\section{Experiments}
\label{sec:experiments}

\subsection{Enterprise agentic Tasks and Experimental Setup}
\label{sec:experimental-setup}

We adopt the task suite, environments, and harness-optimization framework of
\citet{yang2026harnesses}. The suite contains seven objectively verifiable
enterprise tasks spanning payroll auditing, budget approval, stock alerting,
IoT anomaly detection, browser automation, website management, and code
refactoring. We optimize prompts, tools, hooks, context management, and
sub-agent configurations using a GEPA-style search
\citep{agrawal2025gepa}, in which a
\texttt{gemini-3.1-pro-preview} meta-agent proposes failure-driven edits that
are retained only when validation performance improves.

We make two environment changes: agents cannot modify task databases directly
when MCP tools are required, and tool-call errors are exposed to the optimizer.
Consequently, some absolute results may not directly be comparable with those of
\citet{yang2026harnesses}. For model adaptation, we perform LoRA-SFT
\citep{hu2022lora} on Qwen and Gemma using H100/H200 GPUs, converting Gemini
expert trajectories to each student's chat and tool-call format. Training and
validation data are used for optimization, while the test split remains held
out. Unless noted otherwise, we report test success averaged over three runs.
Full model training and inference details are provided in
\textbf{Appendix~\ref{app:model-sft}}.

\subsection{Evolved Harnesses Transfer Upward and Reveal Model-Side Headroom}
\label{sec:results_1}
Consistent with prior reports, evolving the harness lifts task success: mean test success rises from 29.2\% under the base harness to 78.0\% under the evolved harness (Table~\ref{tab:main-v2}, rows 1--2), a gain of 48.8 points that holds across tasks. A task-fitted harness supplies the scaffolding the weaker model needs to succeed reliably, The task-specific edits are detailed in \textbf{Appendix~\ref{app:harness-opt}}. We evolve this harness using only the weaker Qwen model, yet the stronger expert gemini-3.1-pro-preview operates it just as well: the expert improves from 84.4\% under the base harness to 93.6\% under the evolved harness (Table~\ref{tab:main-v2}, rows 3--4, +9.2 on average). A closer look at the trajectories confirms this is real use of the evolved harness components: the expert triggers nearly every evolved edit in 93.6--100\% of its rollouts (Table~\ref{tab:adoption-v2}), including the domain-computation recipe (94.2\%) that the base model rarely uses (30.8\%). Since the expert model demonstrates a superior task success rate on the evolved harness (93.6\% vs.\ 78.0\% on average, +15.6), we could potentially teach the weaker model from the expert to achieve further gains through model updates. This points to a straightforward co-evolution recipe: evolve the harness first, then have the expert model teach the weaker model, pulling up the model arm to maximize the synergy between a stronger model and the evolved harness it stands on.  


\newcommand{\cn}[2]{\makecell{$#1$\\ {\footnotesize$\pm#2$}}}            
\newcommand{\cd}[3]{\makecell{$#1$\\ {\footnotesize$\pm#2$}\\ {\footnotesize$(#3)$}}} 

\begin{table}[t]
  \centering
  \footnotesize
  \setlength{\tabcolsep}{3pt}
  \renewcommand{\arraystretch}{0.95}
  \caption{\textbf{Task success across harness and model adaptations.}
Test-split task success (\%) on seven enterprise benchmarks. Cells report the mean $\pm$ SEM over three runs, with percentage-point changes in parentheses. Changes in rows~2 and~6 are relative to row~1; changes in rows~4, 5, and~7 are relative to row~2. $h_0$ denotes the baseline harness, $h^{*}$ the evolved harness, and \textbf{Avg} the mean across tasks.}

  \label{tab:main-v2}
  \begin{tabular}{clcccccccccc}
    \toprule
    \# & Model & \makecell{Harn.} & Attn & Budget & Stock & Anom & Playwr & Web & Refact & \textbf{Avg} \\
    \midrule
    1 & \makecell[l]{qwen3-coder-\\30b-a3b\\(base)} & $h_0$
      & \cn{13.9}{2.00} & \cn{67.5}{1.03} & \cn{16.7}{1.92} & \cn{0.6}{0.56}
      & \cn{38.4}{2.21} & \cn{0.0}{0.00} & \cn{67.2}{2.00} & \cn{29.2}{0.61} \\
    \midrule
    2 & \makecell[l]{\textbf{qwen3-coder-}\\\textbf{30b-a3b}\\\textbf{(base)}} & $h^{*}$
      & \cd{81.7}{1.11}{+67.8} & \cd{87.6}{0.27}{+20.1} & \cd{76.7}{0.00}{+60.0} & \cd{89.4}{1.11}{+88.8}
      & \cd{97.3}{0.71}{+58.9} & \cd{43.3}{5.09}{+43.3} & \cd{70.0}{4.19}{+2.8}
      & \makecell{$\mathbf{78.0}$\\ {\footnotesize$\pm0.97$}\\ {\footnotesize$(+48.8)$}} \\
    \midrule
    3 & \makecell[l]{gemini-3.1-\\pro-preview} & $h_0$
      & \cn{97.4}{0.93} & \cn{97.2}{0.50} & \cn{51.7}{1.92} & \cn{100.0}{0.00}
      & \cn{88.7}{3.19} & \cn{71.1}{4.44} & \cn{85.0}{0.96} & \cn{84.4}{0.85} \\
    \midrule
    4 & \makecell[l]{gemini-3.1-\\pro-preview} & $h^{*}$
      & \cd{97.0}{1.48}{+15.3} & \cd{93.2}{0.24}{+5.6} & \cd{94.4}{0.56}{+17.7} & \cd{100.0}{0.00}{+10.6}
      & \cd{100.0}{0.00}{+2.7} & \cd{85.6}{1.11}{+42.3} & \cd{85.0}{2.55}{+15.0} & \cd{93.6}{0.46}{+15.6} \\
    \midrule
    5 & \makecell[l]{qwen3-coder-\\30b-a3b\\(SFT-imit.)} & $h^{*}$
      & \cd{51.9}{4.42}{-29.9} & \cd{77.0}{1.85}{-10.6} & \cd{62.6}{0.19}{-14.1} & \cd{85.2}{1.21}{-4.2}
      & \cd{81.3}{1.47}{-16.0} & \cd{23.0}{1.34}{-20.3} & \cd{60.9}{1.82}{-9.1} & \cd{63.1}{0.81}{-14.9} \\
    \midrule
    6 & \makecell[l]{qwen3-coder-\\30b-a3b\\(SFT-imit.)} & $h_0$
      & \cd{27.8}{1.28}{+13.9} & \cd{69.5}{1.82}{+2.0} & \cd{28.3}{2.70}{+11.6} & \cd{21.1}{3.38}{+20.5}
      & \cd{36.8}{5.35}{-1.6} & \cd{0.0}{0.00}{+0.0} & \cd{65.0}{1.67}{-2.2} & \cd{35.5}{1.06}{+6.3} \\
    \midrule
    7 & \makecell[l]{\textbf{qwen3-coder-}\\\textbf{30b-a3b}\\\textbf{(SFT-corr.)}} & $h^{*}$
      & \cd{81.3}{0.49}{-0.4} & \cd{87.0}{1.32}{-0.6} & \cd{78.9}{1.11}{+2.2} & \cd{91.1}{3.38}{+1.7}
      & \cd{98.5}{0.27}{+1.2} & \cd{48.9}{2.22}{+5.6} & \cd{72.2}{1.47}{+2.2}
      & \makecell{$\mathbf{79.7}$\\ {\footnotesize$\pm0.67$}\\ {\footnotesize$(+1.7)$}} \\
    \bottomrule
  \end{tabular}
\end{table}

\subsection{Expert-Trajectory Imitation Degrades Performance Under Evolved Harnesses}
\label{sec:results_2}
To imitate the expert's success on the evolved harness, we LoRA fine-tune the weaker model on the expert's trajectories under that same harness. We collect the expert's successful task completions, convert them to Qwen's input format, and mix them with the weaker model's own successes as training data. Surprisingly, this recipe lowers Qwen's mean test success from 78.0\% to 63.1\% (Table~\ref{tab:main-v2}, rows 2 and 5), and the regression holds across all seven tasks, ranging from $-$4.2 points on anomaly detection to $-$29.9 on payroll auditing ($-$14.9 on average). The largest drops fall on payroll auditing ($-$29.9), website management ($-$20.3), and browser automation ($-$16.0). This is counterintuitive: the teacher is strictly stronger on this harness, and imitating a stronger model is a standard way to transfer capability. The result suggests the failure lies not in the teaching signal itself, but in how it interacts with the evolved harness.

\newcolumntype{E}{>{\centering\arraybackslash}p{0.63in}}

\begin{table}[t]
  \centering
  \footnotesize
  \setlength{\tabcolsep}{2.5pt}
  \renewcommand{\arraystretch}{1.05}
 \caption{\textbf{Adoption of evolved harness edits across models.}
Each entry reports the percentage of rollouts in which the corresponding harness edit is used at least once. All models operate under the same harness $h^{*}$ evolved for Qwen.}
  \label{tab:adoption-v2}
  \begin{tabular}{l EEEEEE}
    \toprule
    & \textbf{Create tool} & \makecell{\textbf{Tool-use}\\\textbf{example}} & \makecell{\textbf{Static}\\\textbf{reinforce.}}
    & \makecell{\textbf{Enforce}\\\textbf{via hook}} & \makecell{\textbf{Env/API}\\\textbf{convention}} & \makecell{\textbf{Domain}\\\textbf{computation}} \\
    \midrule
    \emph{Description}
      & {\scriptsize Create a tool the base seed lacked}
      & {\scriptsize Teach a tool-use convention / worked example}
      & {\scriptsize Reinforce a prompt instruction}
      & {\scriptsize Block a forbidden action with a hook}
      & {\scriptsize State an environment / API convention}
      & {\scriptsize State a multi-step arithmetic recipe} \\
    \midrule
    gemini-3.1-pro-preview & 99.3 & 93.6 & 100.0 & 99.0 & 99.1 & 94.2 \\
    \makecell[l]{qwen3-coder-30b-a3b\\ {\scriptsize(base)}} & 98.7 & 93.5 & 96.4 & 100.0 & 98.2 & \textbf{30.8} \\
    \makecell[l]{qwen3-coder-30b-a3b\\ {\scriptsize(SFT-imitation)}} & 91.2 & 89.6 & 91.7 & 99.0 & 97.5 & \textbf{76.1} \\
    \bottomrule
  \end{tabular}
\end{table}

\subsubsection{Ablation I: Expert Imitation Can Help on Baseline Harnesses}
To test whether the regression comes from imitation itself, we apply the identical recipe under the un-evolved baseline harness. Here it helps: fine-tuning on the expert's successful-completion trajectories raises Qwen's mean test success from 29.2\% to 35.5\% (Table~\ref{tab:main-v2}, rows 1 and 6, +6.3 on average). The same expert trajectories that degrade the model under the evolved harness, instead improve task success through model SFT under the baseline harness. The teaching signal is therefore not the problem in itself; the regression is specific to the evolved harness, which points to an interaction between imitation and harness evolution as the cause.

\subsubsection{Ablation II: Similar regression persists for a Gemma model}
To rule out that the regression is an artifact of a specific weak--strong model pairing (Qwen--Gemini), we repeat the harness-first, expert-imitation fine-tuning workflow with a different weak student, gemma-4-26b-a4b-it, on the Webarena task. The evolved harness lifts the weak Gemma from 46.7\% to 55.6\% (+8.9). It also transfers upward to the expert: Gemini gains from 71.1\% to 81.1\% (+10.0) on the same harness, just as we saw with Qwen. Yet fine-tuning the weak Gemma on the expert's trajectories again regresses it, to 41.1\%. That is 14.5 points below its evolved-harness baseline, and 5.6 points below even its default-harness baseline. Gemma is a reasoning model whose native style may more closely match the Gemini expert's, yet the regression still persists. The interaction between imitation and harness evolution, rather than the model pairing or output style, is therefore the more likely cause.

\subsubsection{Analysis of Imitation-Induced Degradation}
We first examine whether the fine-tuned models reduce the use of the evolved harness after imitating the expert. To the contrary, the model still exercises the harness's task-specific components: it triggers almost every harness edit in its rollouts, and it actually raises its use of the domain-computation recipe from 30.8\% to 76.1\% (Table \ref{tab:adoption-v2}). So the fine-tuned model uses the evolved scaffold more, not less, and it demonstrably acquires the domain knowledge that recipe expert encodes.

To locate what breaks, we classify every failed rollout against the full six-category adaptation-failure ontology of \citet{yang2026harnesses} and compare the failure distribution before and after imitation fine-tuning (detailed analysis methods in Appendix~\ref{app:analysis-method}). The change is concentrated in two of the six categories; the rest (tool-use, instruction-following, long-context, other) shift only marginally. The first is implicit-knowledge failures, where the agent misses implied constraints or conventions, skips steps a domain practitioner would take automatically, or fails to apply domain-specific heuristics. The second is planning failures, where the agent omits critical subtasks, commits to a wrong plan without recovery, or spins in replanning loops that make no forward progress. Table~\ref{tab:fail-comp} shows the split, and the two categories move in opposite directions. On the knowledge axis, expert-imitation fine-tuning helps: the implicit-knowledge bucket shrinks from 46.2\% to 44.5\% of failures ($-$1.7) rather than growing (Table~\ref{tab:adoption-v2} supplements this, confirming the model now applies the domain-computation recipe far more often, 30.8\% to 76.1\%). What breaks instead is planning: planning failures emerge as an essentially new failure mode, rising from 1.1\% to 14.6\% of failures (+13.5). This suggests that imitation successfully transfers the expert's knowledge, but also transfers the expert's planning style, which the weaker model cannot follow faithfully after a lightweight LoRA fine-tune. As a result, it no longer matches the harness that was evolved around its own native planning behavior. For example, on a payroll-audit task the fine-tuned model works out the correct answer but never manages to submit it: having lost its own step-by-step rhythm, it keeps re-checking its work instead of finishing, so the run never completes and scores zero despite having the right result in hand (Appendix~\ref{app:case-studies}). This also explains why the same recipe helps on the baseline harness but hurts on the evolved one. On the baseline harness, imitation induces the same planning failures (they rise from 0.9\% to 11.5\% of failures, +10.6), but that harness was never fit to the model's native planning in the first place, so there is no planning fit to lose; the knowledge gain therefore dominates (the knowledge bucket falls from 61.0\% to 54.4\%, $-$6.6), and net success rises. The evolved harness, by contrast, earned its gains through the model's native planning behavior, so disrupting that fit erases the improvement. The failure is therefore not in the teaching signal itself, but in the loss of model--harness fit it induces.

\newcolumntype{K}{>{\centering\arraybackslash}p{0.80in}}
\newcolumntype{L}{>{\raggedright\arraybackslash}p{1.02in}}

\begin{table}[t]
  \centering
  \footnotesize
  \setlength{\tabcolsep}{2.5pt}
  \renewcommand{\arraystretch}{1.1}
  \caption{\textbf{Composition of hard failures before and after model adaptation.}
Entries report the percentage of each condition's hard failures attributed to lack of knowledge or planning defects. Parentheses show the percentage-point change from the base model under the same harness. $h_0$ denotes the baseline harness and $h^{*}$ the evolved harness. Detailed analysis methods are described in Appendix~\ref{app:analysis-method}.}
  \label{tab:fail-comp}
  \begin{tabular}{L KKKKK}
    \toprule
    \textbf{Failure type}
      & \makecell{(base)\\ $h_0$}
      & \makecell{(SFT-imit.)\\ $h_0$}
      & \makecell{(base)\\ $h^{*}$}
      & \makecell{(SFT-imit.)\\ $h^{*}$}
      & \makecell{(SFT-corr.)\\ $h^{*}$} \\
    \midrule
    \textbf{Lack of knowledge}
      & 61.0 & 54.4~{\scriptsize($-6.6$)} & 46.2 & 44.5~{\scriptsize($-1.7$)} & 43.2~{\scriptsize($-3.0$)} \\
    \textbf{Planning defects}
      & 0.9 & 11.5~{\scriptsize($+10.6$)} & 1.1 & 14.6~{\scriptsize($+13.5$)} & 1.8~{\scriptsize($+0.7$)} \\
    \bottomrule
  \end{tabular}
\end{table}

\subsection{On-Policy Expert Correction Makes Harness and Model Adaptation Compose}
\label{sec:results_3}

To introduce the expert's teaching signal without breaking planning fit, we instead correct the weaker model on-policy rather than imitating entire expert trajectories. We design a data-synthesis pipeline, driven end-to-end by a self-directed MLE agent, that preserves the weaker model's own planning distribution. Starting from the weaker model's own rollouts under the evolved harness, the agent localizes the single turn at which each failed rollout goes wrong, and an expert model then rewrites only that one turn in place. In this way we introduce only the necessary correction, leaving every surrounding step the model produced untouched. We then LoRA-SFT the weaker model on this set of minimally edited trajectories and evaluate it under the same evolved harness.

Our results show that this makes the two levers---harness evolution and model fine-tuning---compose. On-policy correction matches or beats the evolved-harness base on every task (Table~\ref{tab:main-v2}, row 7): it raises mean test success from 78.0\% to 79.7\% (+1.7), gaining on five of seven tasks---website management (+5.6), stock alerting (+2.2), code refactoring (+2.2), anomaly detection (+1.7), and browser automation (+1.2)---and staying within noise on the two tasks where the harness has likely already saturated (budget approval $-$0.6, payroll auditing $-$0.4). This contrasts sharply with direct imitation, which regressed on all seven tasks ($-$14.9 on average): correcting the model on its own trajectories adds capability where the harness left headroom and preserves it where the harness was already near ceiling.

The failure analysis shows why the gain is safe (Table~\ref{tab:fail-comp}). Under on-policy correction, the planning bucket stays at the base-model floor, moving only from 1.1\% to 1.8\% of failures (+0.7)---in sharp contrast to imitation's jump to 14.6\%---so the model--harness fit is preserved. At the same time the knowledge bucket falls further, from 46.2\% to 43.2\% ($-$3.0), and the overall failure rate drops as well (28.9\% to 26.8\%). On-policy correction therefore delivers the same knowledge improvement that imitation did, but without the planning collapse that breaks the fit between the weaker model and the evolved harness. It does not fully close the gap between the weaker and the expert model, likely due to a limit of the lightweight LoRA approach, but it offers a principled and low-cost way to maximize the joint synergy between an evolved harness and model training, without introducing counterproductive regression. Because the correction is generated entirely by the self-directed MLE agent, it can be safely stacked into an iterative co-evolution of both harness and model.


\section{Conclusion}
\label{sec:conclusion}
Solving complex, long-horizon enterprise tasks reliably is challenging, and an optimized harness has shown promise in helping a weaker model do so at a fraction of the cost of a stronger frontier model. In this study we asked whether co-optimizing the model weights alongside harness evolution, through lightweight fine-tuning, can push performance further still. We first find that even though the harness is evolved for a weaker model, a stronger expert model uses it well, which suggests the expert could show the weaker model how to close the gap. However, under the evolved harness, simply imitating the expert's trajectories regresses the weaker model ($-$14.9 on average). Our analysis shows why: imitation shifts the model's planning style, which disrupts its fit with the harness and cancels out the harness's benefits. We then develop a self-directed MLE pipeline for on-policy correction, which synthesizes training data and fine-tunes the model on its own rollouts. This approach delivers a further gain without eroding the gains already achieved through harness evolution, and offers a sound way to co-evolve the model and the harness. The recipe is also lightweight and economical: training takes under an hour, so it can be stacked safely into a co-evolution loop. More broadly, our results show that staying on-policy is what matters in the harness--model setting: the teaching signal is only useful when it respects the fit between the model and the harness it operates on.

Co-evolution of both models and harnesses is an exciting direction. In future work, we plan to combine reinforcement learning with on-policy expert guidance to push the weaker model further where headroom remains. We also want to make harness evolution aware that the model will later be fine-tuned, so the two arms can be jointly optimized rather than run in sequence.

\bibliographystyle{plainnat}
\bibliography{references}


\clearpage
\appendix
\section*{Appendix}
\label{app:appendix}

\section{Harness Optimization}
\label{app:harness-opt}

We evolve each task's harness using codebase from ~\citep{yang2026harnesses}, with the weaker model
(qwen3-coder-30b-a3b or gemma4-26B-A4B) as the executor and \texttt{gemini-3.1-pro-preview} as the
reflection model. The editable component is the harness itself---primarily the agent's
system prompt, but the proposer is also free to add or remove tools and hooks. For each task, every optimization iteration reflects on a minibatch, proposes a harness edit, and retains the edit only if it improves performance on that minibatch; retained candidates are then added to the Pareto pool. We run each task under three random seeds with a budget
of \$20 and a 600\,s per-rollout time limit, and select the best harness $h^{*}$ as the
candidate with the highest validation score (ties broken by the earliest iteration). Table~\ref{tab:harness-opt} summarizes, per
task, the failure category the winning edits address and the substance of those edits.

\begin{table}[H]
  \centering
  \footnotesize
  \setlength{\tabcolsep}{4pt}
  \renewcommand{\arraystretch}{1.25}
  \caption{Dominant harness adaptation per task (winning seed). \emph{Cat.}\ is the
  failure category the winning edits address, using the ontology of
  Appendix~\ref{app:analysis-method}: 1a/1c create tools / add tool-use instructions,
  2a/2c/2d reinforce an instruction via prompt / tool / hook, 3a externalize implicit
  knowledge as instructions, 5a/5b add an explicit plan. \emph{Harness adaptation}
  describes the substance of the winning edits.}
  \label{tab:harness-opt}
  \begin{tabular}{l l >{\raggedright\arraybackslash}p{3.3in}}
    \toprule
    Task & Cat. & Harness adaptation \\
    \midrule
    Attendance & 3a, 1c &
      Spell out the aggregation and payroll rules the base model got wrong (per-employee
      averaging, the overtime policy). Add workarounds for files the editor cannot open. \\
    Anomaly & 3a, 1c &
      A single rewrite lays out the full query-detect-report-upload workflow, including
      actually calling the upload tool. This alone saturates the validation set. \\
    Stock & 1a, 3a, 2c &
      Add a tool to enumerate the full catalog and require that every low-stock item be
      processed, not a subset. A residual under-enumeration ceiling remains. \\
    Playwright & 3a &
      One rewrite states the sandbox conventions a generic agent cannot infer (run
      headless, use local files, write the exact number of tests) and to verify locally
      before finishing. \\
    Website & 1a, 3a &
      Add the structured finish tool the harness lacked, then externalize the admin-console
      conventions needed to read the correct numbers. \\
    Budget & 3a, 5a/5b &
      Externalize the core protocol (initiate contact; never read backend state directly),
      then add structural fixes: an explicit ordered plan and a state-access guard. \\
    Refactor & --- &
      Insert a global search tool. \\
    \bottomrule
  \end{tabular}
\end{table}

\section{Model SFT}
\label{app:model-sft}

\paragraph{Setup.} All fine-tuning recipes use the same base model and LoRA
configuration. We fine-tune \texttt{Qwen/Qwen3-Coder-30B-A3B-Instruct} with LoRA (rank
16 or rank 64 and report the best scores) for 2 epochs at learning rate
$1\mathrm{e}{-}4$ in bf16, with an effective batch size of 8. We train at a sequence length
of 49\,152 tokens and additionally on a longer 98\,304-token context if > 5 \% training data suffers from truncation. Trained adapters are merged into a dense checkpoint
($\sim$57--61\,GB) and served with tensor parallelism across two H200 GPUs; a single run
completes in under an hour. The Gemma replication (Appendix~\ref{app:analysis-method})
fine-tunes \texttt{gemma-4-26b-a4b-it} with the analogous LoRA recipe.

\paragraph{Imitation data (SFT-imit.).} We collect trajectories that score $1.0$ under
each task's evolved harness $h^{*}$---first from the \texttt{gemini-3.1-pro-preview}
expert, then from the base model's own passing rollouts---convert them to the model's
chat format, and fine-tune directly on them. For the Gemma replication on the Webarena
task we also shape the gemini-3.1-pro-preview trajectories to fit Gemma model's reasoning format. The
headline imitation arm is reported as the mean over three independent LoRA-training seeds
(0, 101, 202).

\paragraph{On-policy correction data (SFT-corr.).} Rather than the expert's trajectories,
we start from the base model's \emph{own} rollouts under $h^{*}$ and build a small
($\sim$~ 500-row) training set from in-place expert correction
($\sim$400 rows), where an automated failure-locus step identifies the
single turn at which a failed rollout goes wrong and an expert model rewrites only that
turn, keeping every surrounding step unchanged. We mixed in ~self-pass ($\sim$50 rows) sampled
from the model's own passing rollouts, prioritizing capability-boundary examples that
pass on some rollouts but not all. 

\paragraph{Correction prompt and best-of-$N$ selection.} For each localized failing turn,
the expert model is given the task instructions, the model's trajectory up to
that turn, and the specific checkpoint that failed, and is asked to rewrite \emph{only}
that turn---one sentence of corrected strategy followed by the corrected tool call---while
leaving all preceding and following steps intact. We sample $N=3$ candidate corrections
per turn and keep the best one under a quality judge. 

\section{Analysis Methodology}
\label{app:analysis-method}

\paragraph{Data.} The failure-composition analysis in Table~\ref{tab:fail-comp} operates
on failed rollouts. Task success is binary, so a hard failure is a rollout that scores
$0$. For each condition we
collect all of its failed rollouts under the relevant harness and classify them. 

\paragraph{Method.} We classify each failed rollout into a six-category adaptation-failure
ontology (categories 0--5): 0~other, 1~tool-use, 2~instruction-following, 3~implicit
knowledge, 4~long-context, and 5~planning. The two categories the model lever moves are
implicit knowledge---the agent finishes with a wrong domain value, misses an implied
convention, or skips a step a practitioner would take automatically---and planning---the
agent omits a critical subtask, commits to a wrong plan without recovery, or loops in
replanning without progress. Labeling is done by a LLM-as-judge classifier over each rollout's failed
checkpoints and tool errors. We report each category both as a share of that condition's
failures and as a share of all rollouts. In a companion check we run the same classifier
as an adoption scan---measuring, from the trajectories, how often each model actually
invokes each evolved-harness component (Table~\ref{tab:adoption-v2}).

\section{Case Studies}
\label{app:case-studies}

We give two rollouts that show how expert imitation breaks the fit between the model and
the evolved harness. In both, the fine-tuned model has the domain knowledge it needs; what
it loses is the planning cadence the harness was tuned around.

\paragraph{Case 1: payroll audit (Qwen).} The evolved harness for this task supplies an
explicit compute recipe (employee-level aggregation first, then a department-level
mean-of-means; ceiling-then-multiply payroll; fixed rounding) and a single
output-verification step: print the final columns and confirm they match the requested
schema before saving. Under the base model this scaffold works cleanly---on one example
all three rollouts score $1.0$ in 14--20 steps, because the harness's ``compute, verify
once, finish'' cadence was fit to the model's own step-by-step planning. After imitation
fine-tuning the model still writes the correct output (the scorer even notes the correct
department aggregates), but with its planning cadence stripped it turns the single verify
step into an unbounded loop: it re-issues the same view 29 times, re-reads the prompt
about 20 times, accumulates tool errors, and never emits a finish across 78 steps, so the
rollout scores $0$. Same knowledge, same harness---only the post-imitation plan style
differs, and it no longer lands on the finish the scaffold expects.

\enlargethispage{2\baselineskip}
\paragraph{Case 2: Webarena review count (Gemma).} The evolved harness here adds a
``Magento Admin Conventions'' block whose load-bearing piece is a grid-filter procedure:
locate the status column, apply the filter, then read the filtered total, and return it
through a structured finish tool. For ``total count of Approved reviews'' (truth 346), the
base model follows this ``filter, then count'' procedure and answers correctly. The
imitation-fine-tuned Gemma copies the expert's terse, answer-early style but not the
competence behind it: it navigates to the review grid but never forms the ``filter, then
count'' sub-goal, reads the unfiltered ``351 records found'' total straight off the grid,
and finishes with 351. The same reflex recurs on other review questions where the
unfiltered total cannot be right. Because Gemma already shares the expert's terse output
shape, this case shows the problem is not surface style but a confidence and brevity the
weaker model cannot back up---and it bypasses exactly the multi-step procedure the harness
relied on.

\end{document}